\documentclass[conference]{IEEEtran}
\usepackage{graphicx}
\usepackage{subfigure}
\usepackage{amsmath}
\usepackage{amssymb}
\usepackage{url}

\begin{document}

\title{Exploring Energy-Accuracy Tradeoffs in AI Hardware}

\author{\IEEEauthorblockN{Cory Merkel}
\IEEEauthorblockA{Brain Lab\\Department of Computer Engineering\\
Rochester Institute of Technology\\
Rochester, New York 14623\\
Email: cemeec@rit.edu}}

\maketitle

\makeatletter



\begin{abstract}
Artificial intelligence (AI) is playing an increasingly significant role in our everyday lives.  This trend is expected to continue, especially with recent pushes to move more AI to the edge.  However, one of the biggest challenges associated with AI on edge devices (mobile phones, unmanned vehicles, sensors, etc.) is their associated size, weight, and power constraints.  In this work, we consider the scenario where an AI system may need to operate at less-than-maximum accuracy in order to meet application-dependent energy requirements.  We propose a simple function that divides the cost of using an AI system into the cost of the decision making process and the cost of decision execution.  For simple binary decision problems with convolutional neural networks, it is shown that minimizing the cost corresponds to using fewer than the maximum number of resources (e.g. convolutional neural network layers and filters).  Finally, it is shown that the cost associated with energy can be significantly reduced by leveraging high-confidence predictions made in lower-level layers of the network.
\end{abstract}

\section{Introduction}

The field of artificial intelligence (AI) is being reinvigorated by the successes of deep learning (DL).  In simple terms, DL algorithms model a particular function (e.g. object identify) by learning more and more abstract representations or features of inputs.  For example, in an image classification task, a DL algorithm may extract edges, corners, shapes, etc. from the pixels of an image.  From there, the algorithm may recognize doors, mirrors, and eventually make and model for an automobile identification task.  Indeed, it is well-known that human vision also makes use of gradual feature abstraction, with low-level features being extracted from the retina, and more abstract features being recognized in the ventral pathway from the V1 to the IT cortices \cite{kandel2000principles}.  While core object recognition in humans takes approximately 100 ms, it has recently been shown that humans can recognize objects much faster ($\approx$13 ms) \cite{potter2014detecting}, albeit at lower accuracy.  This shows how the brain can make quick, low-energy predictions based on unrefined information and then refine those predictions based on further processing.  This paper explores to what degree we can apply similar principles in DL for applications that have extreme size, weight, and power constraints.


\begin{figure}[!t]
    \centering
    \includegraphics[width=\columnwidth]{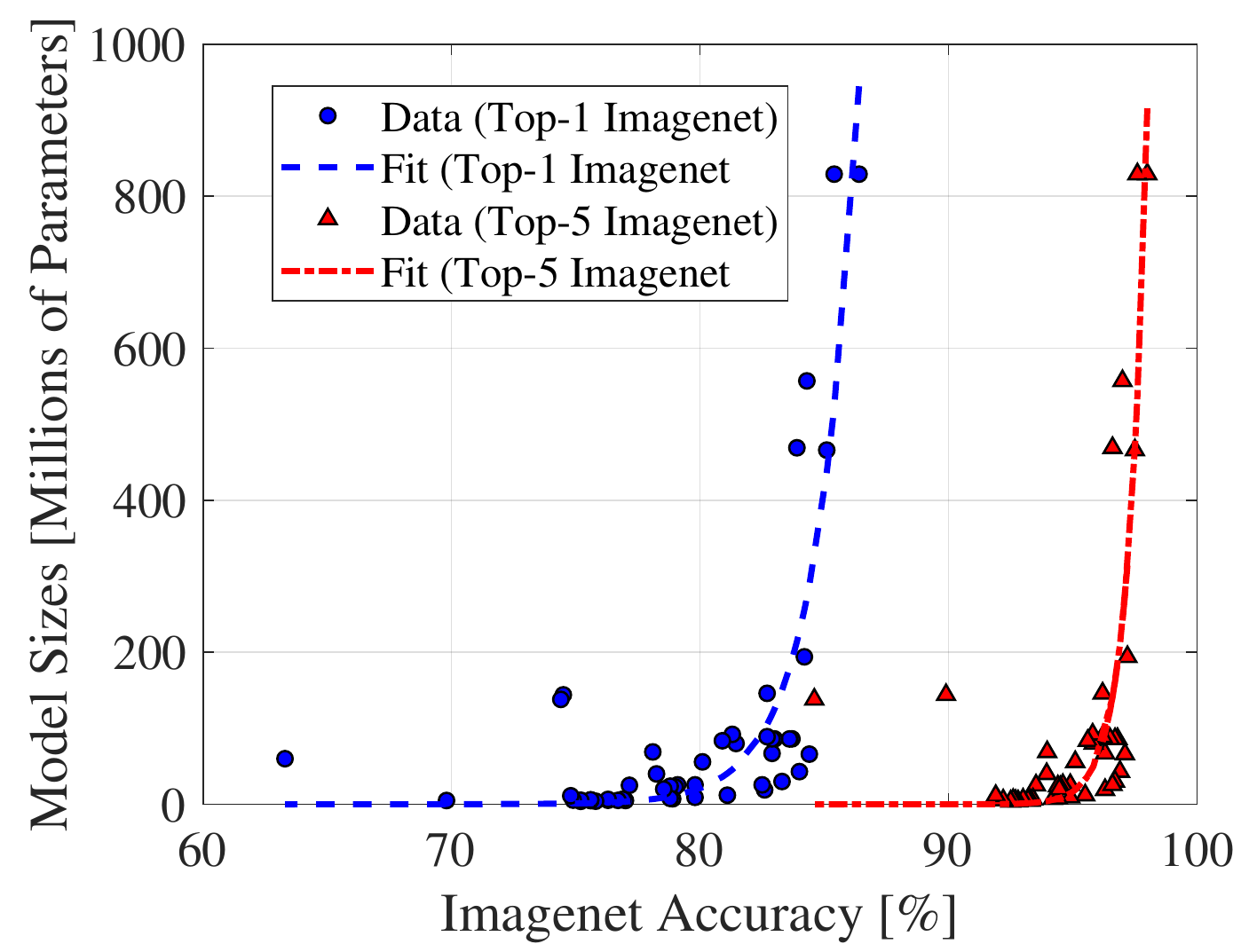}
    \caption{ANN model size vs. accuracy on the ImageNet dataset.}
    \label{fig:cnn_imagenet}
\end{figure}

The number of parameters in state-of-the-art DL algorithms is typically on the order of hundreds of millions, with some even surpassing the one billion mark.  Figure \ref{fig:cnn_imagenet} shows the model size in millions of parameters vs. accuracy on the ImageNet \cite{imagenet} dataset for several published artificial neural networks (ANNs) \cite{paperswithcode}.  The blue circle datapoints are the top-1 testing accuracies, which are the percentage of times each model's top prediction was correct.  The dashed trendline is a simple exponential fit.  The result is that modest gains in model performance require exponentially larger ANNs.  The same trend is observed in the ImageNet top-5 accuracies as well.   Several methods are being proposed to reduce the size of modern DL models by compressing ANNs via quantization/pruning and sharing techniques, low-rank factorization methods, transferred/compact weight methods, and knowledge distillation, among others \cite{cheng2017survey}.  While these compression methods enable the implementation of DL algorithms on resource-constrained platforms, they do not facilitate dynamic scaling of energy.  In contrast, there have been some recent works that explore runtime energy-accuracy tradeoffs in DL models.  In \cite{tann2016runtime}, the authors explore the dynamic configuration of the number of convolutional neural network (CNN) channels in order to reduce energy with a minimal effect on the overall inference accuracy.  Reference \cite{kim2016dynamic} exploits the relationship between bit stream length and accuracy in stochastic DL implementations to enable dynamic energy-accuracy scaling.  Other works such as \cite{pagliari2018dynamic} explore adjusting the bit width of the DL algorithm in order to reduce energy according to application constraints.  This paper builds on previous work to further explore the energy-accuracy tradeoffs in AI hardware.  Specifically, the contributions of this work are:
\begin{itemize}
    \item A model that quantifies the cost associated with an AI's decision making process and decision execution process
    \item Analysis of the optimal number of CNN layers in a binary decision making problem
    \item A method of reducing CNN energy by dynamically configuring the number of layers based on prediction confidence
\end{itemize}

The rest of this paper is organized as follows: Section \ref{section:cost} defines a cost associated with using an AI system to make decisions and presents a strategy for basing that cost on the AI's energy usage.  Section \ref{section:minimize} discusses minimization of an AI system's cost by operating at an optimal energy-accuracy point.  The case where the cost changes dynamically is discussed in Section \ref{section:dynamic}.  Further cost reduction by using prediction confidence and a dynamic number of layers is presented in Section \ref{section:reducing}.  Finally, Section \ref{section:conclusions} concludes this work.

\section{Neural Network Inference Cost}
\label{section:cost}


The tradeoff between neural network size and accuracy can be generalized by defining a cost function associated with using any AI system in a decision making process:
\begin{equation}
C=\alpha C_{D}+(1-\alpha)C_{X},
\label{eqn:cost}
\end{equation}

\begin{figure*}[t]
\subfigure[]{
\centering
\includegraphics[width=\columnwidth]{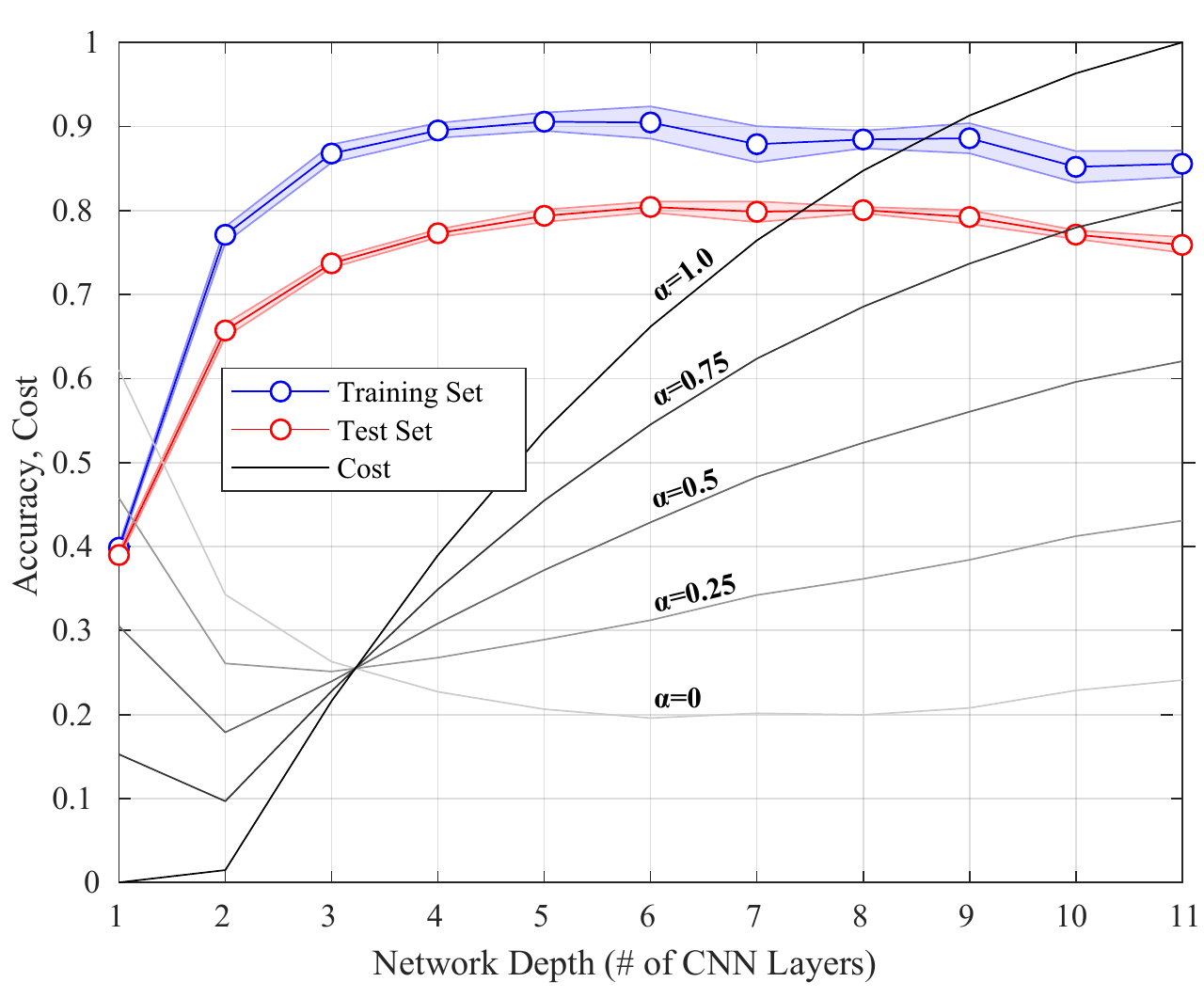}
}
\subfigure[]{
\centering
\includegraphics[width=\columnwidth]{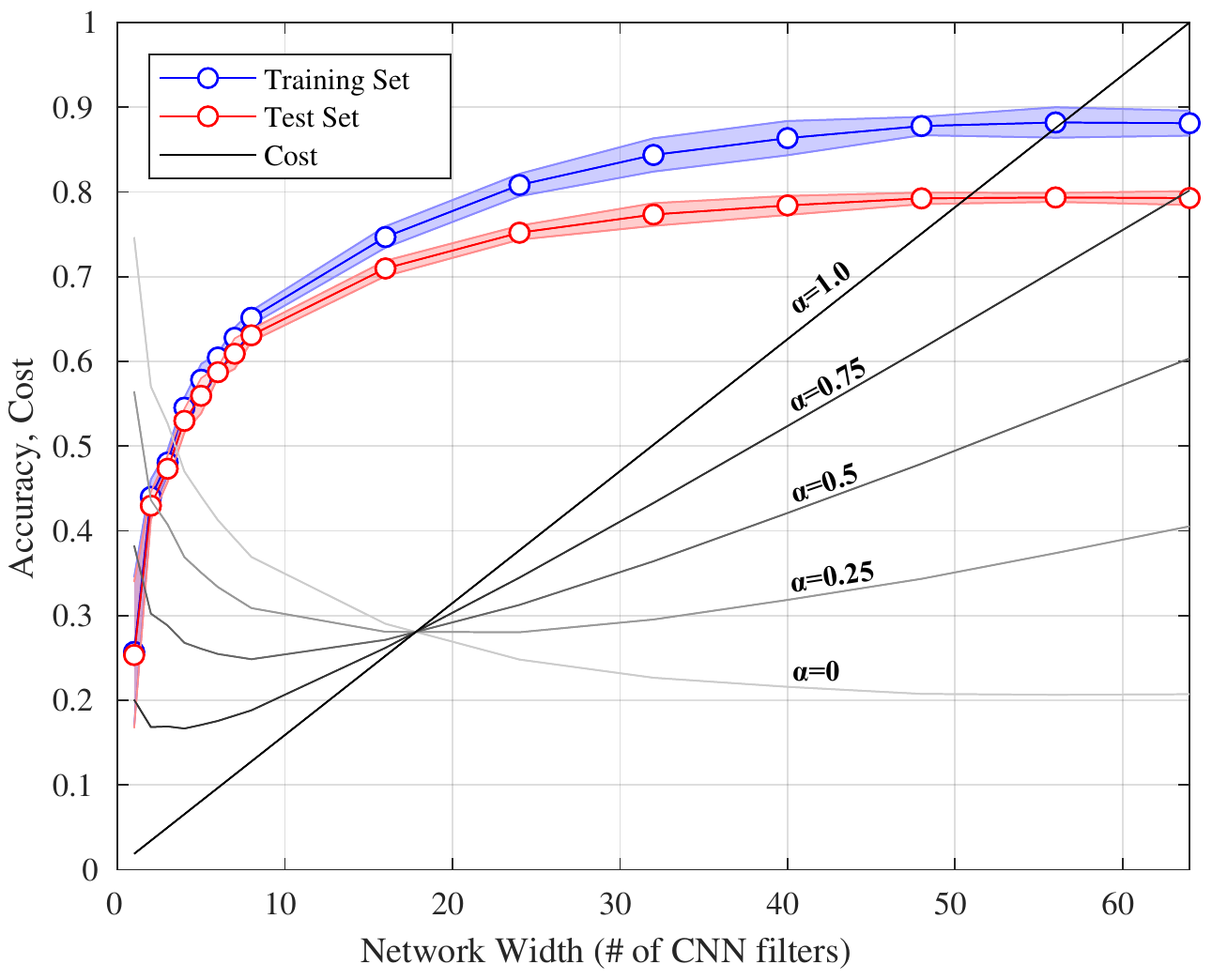}
}
\caption{Accuracy and cost vs. number of layers for a CNN trained on the CIFAR-10 classification dataset.}
\label{fig:baseline}
\end{figure*}
\noindent
where $C_{D}$ and $C_{X}$ are the costs of making and executing a decision based on the AI's computation(s), respectively, normalized between 0 and 1.    Specifically, this paper focuses on the tradeoff between energy and accuracy in ANNs, so $C_{D}$ and $C_{X}$ will be related to the energy associated with ANN inference at a given level of accuracy.  The value $\alpha$ is between 0 and 1 and assigns a relative importance to the two cost factors.  Here,  $\alpha=0$ corresponds to the case where the cost of executing a bad decision is considered much worse than the cost of making that decision.  In contrast $\alpha=1$ means that the system should spend as little energy on making a decision as possible, even if it is a bad decision.  Roughly, one may associate smaller $\alpha$ values with cloud computing and larger $\alpha$ values with edge computing.  The calculation of $\alpha$, $C_{D}$, and $C_{X}$ are application and implementation dependent.  In this work, we are interested in energy costs, so 
\begin{equation}
C_{D}\equiv\frac{E_{D}}{E_{D_{max}}},
\end{equation}
\noindent
where $E_{D}$ is the energy required for the ANN to perform inference, and $E_{D_{max}}$ is the maximum energy required to perform inference within the space of ANNs being compared.  The energy will be implementation dependent, but here we assume that it is proportional to the number of multiply-accumulate (MAC) operations.  For a convolutional layer in a CNN, this can be estimated as
\begin{equation}
MAC_{conv}\approx (W_{i}-W_{f}+1)(H_{i}-H_{f}+1)N_{c}W_{f}H_{f}N_{f},
\end{equation}
where $W_{i}$, $W_{f}$, $H_{i}$, and $H_{f}$ are the width and height of the layer input ($i$) and convolutional filter ($f$), respectively, $N_{c}$ is the number of input channels, and $N_{f}$ is the number of filters.  For a fully-connected layer, the MACs are estimated as
\begin{equation}
MAC_{fc}\approx N_{u}N_{y},
\end{equation}
where $N_{u}$ and $N_{y}$ are the number of layer inputs and outputs, respectively.  

Now, consider an $L$-layer CNN with $L-1$ convolutional layers and 1 fully-connected output layer.  If the cost of executing a decision based on the CNN's classification is $1-A$ (where $A$ is the accuracy), then the overall cost becomes
\begin{equation}
\begin{split}
C&=\alpha\frac{E_{D}}{E_{D_{max}}}+(1-\alpha)(1-A)\\
&=\alpha\frac{\sum\limits_{i=1}^{L-1} MAC_{conv_{i}}+MAC_{fc}}{MAC_{max}}+(1-\alpha)(1-A)
\end{split}
\end{equation}
where $MAC_{max}$ is a normalization constant that is equal to the maximum number of MACs in any CNN within the design space.  For example, if comparing the costs of all CNNs with $L$ ranging from 1 to 11, $MAC_{max}$ would be the number of MACs for the CNN with 10 convolutional layers and one fully-connected output layer.  In this paper, each of the $L-1$ CNN convolutional layers has the same width and uses 3x3 filters.  The fully-connected 10-unit output layer employs a softmax activation function.  For all experiments, the network is trained on the CIFAR-10 dataset using the Adam optimizer with dropout applied to the fully-connected layer (rate = 0.8), $\ell_{2}$ regularization on all of the weights (magnitude = 0.001), and early stopping.  Figure \ref{fig:baseline}(a) shows the cost as well as the training and test set accuracies for CNNs of this type with the number of layers $L$ (network depth) ranging from 1 to 11, and 64 filters for each convolutional layer.  The best performance is achieved with 6 layers (5 convolutional and 1 softmax).  Beyond that, increasing the number of layers causes overfitting, reducing the test accuracy.  Note that state-of-the art CNNs can achieve close to 100\% test accuracy on the CIFAR-10 dataset using a number of architectural and training enhancements such as dataset augmentation, skip connections, etc.  However, the goal in this work is to explore the relative tradeoffs between energy cost and network accuracy, not to achieve state-of-the-art performance.  When $\alpha=0$, the lowest cost occurs with a 6-layer network, which is the one with the best test accuracy.  Figure \ref{fig:baseline}(b) shows the accuracy and cost vs. network width with the depth fixed at 6 layers.  Note that in this case, the normalization constant for the decision cost is the number of MACs required for a 6-layer network with 64 filters in each convolutional layer.  In both cases--varying network depth and varying network width--a non-zero $\alpha$ value implies that the minimum cost may be achieved by using fewer than the maximum available resources (e.g. layers or filters).

\section{Minimizing the Cost}
\label{section:minimize}

\begin{figure}[!t]
    \centering
    \includegraphics[width=\columnwidth]{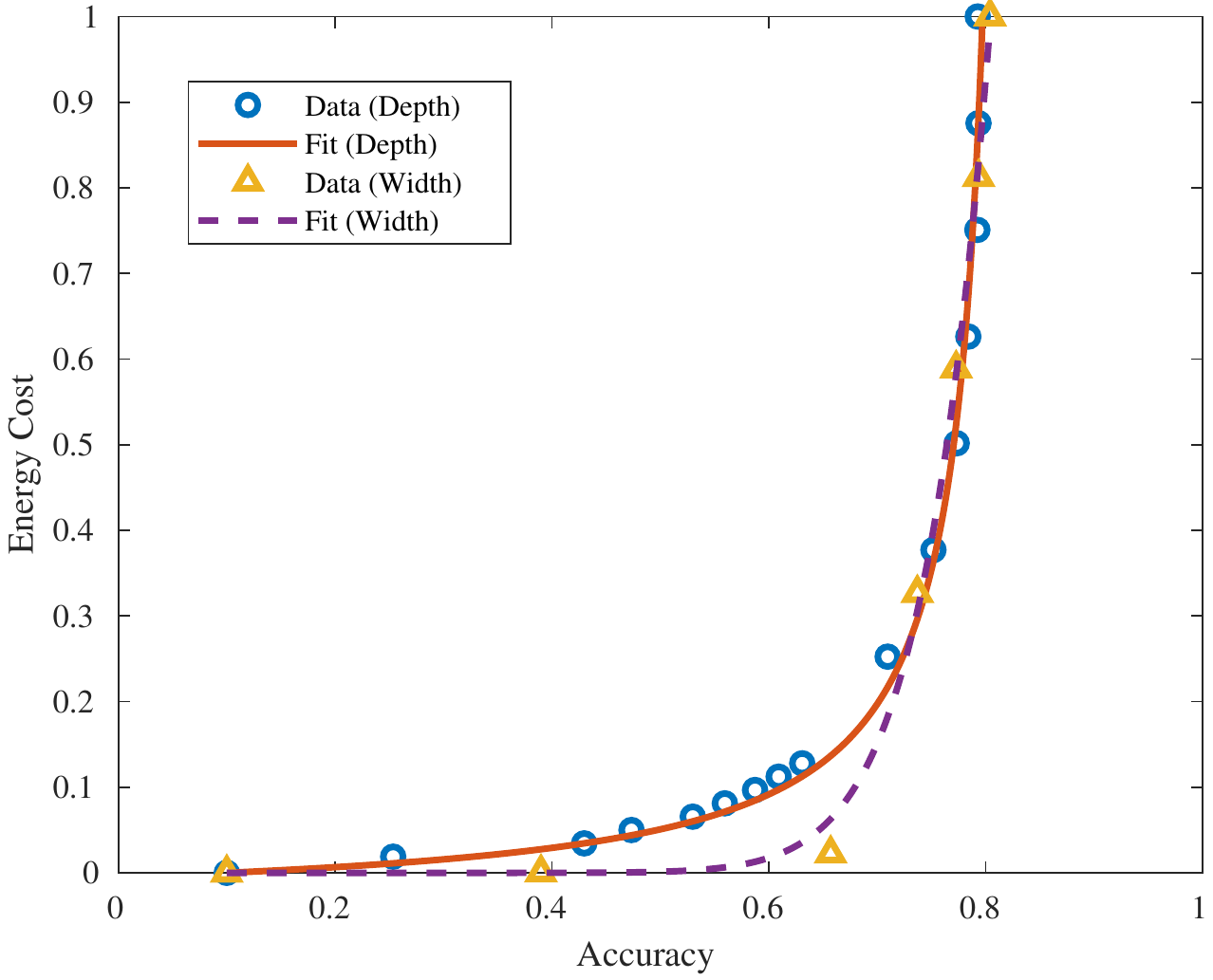}
    \caption{Energy-accuracy curves for a CNN with varying width and depth.}
    \label{fig:energy_accuracy}
\end{figure}

The previous section showed that the relative importance of the decision and execution costs dictates the optimal level of computational resource usage in an ANN.  This corresponds to an operating point on the network's energy-accuracy curve.  Figure \ref{fig:energy_accuracy} shows the energy-accuracy curves for the CNN architecture described in the last section with varying depth (up to 6 layers) and varying width (up to 64 convolutional filters).  The curve corresponding to varying depth is fit to an exponential function, $C_{D}=ae^{bA}$, while the curve corresponding to varying width is better fit with a rational function, $C_{D}=a(A-0.1)/(1-bA)$, where $a$ and $b$ are fitting parameters.  The remainder of this paper focuses on variations of the network depth with the width fixed at 64 convolutional filters.  The optimal depth can be found by minimizing the cost function:

\begin{equation}
\frac{\partial C}{\partial A}=\alpha \frac{\partial}{\partial A}C_{D}(A)+\frac{\partial}{\partial A}(1-\alpha)(1-A)=0.
\label{eqn:dcostA}
\end{equation}

\noindent
Plugging in the exponential for for $C_{D}$, the accuracy at the optimal energy-accuracy cost is given by:
\begin{equation}
A^{*}=\frac{1}{b}\mathrm{log}\left(\frac{1-\alpha}{\alpha ab}\right).
\label{eqn:acccostmin}
\end{equation}
\noindent
When $\alpha=1$, the cost is minimized by using the fewest computational resources.  Notice that the accuracy which minimizes the overall cost will always have this form when $C_{X}$ is a linear function of accuracy (as long as $C_{D}$ also maintains its exponential form). 
In the previous section, we showed that the optimal energy-accuracy point for an AI system depends on the relative cost of the energy to make a decision and the potential cost of making a bad decision.  In general, this relative cost factor $\alpha$ is very difficult to quantify since the consequences of bad decisions can be very complex.  For example, bad decisions in defense applications may affect national security, and bad decisions in medical applications could cost human life.  Furthermore, the energy cost associated with decision making could involve much more than just MAC operations.  In general, decision making may involve time and effort of humans and may also have a monetary cost.

To illustrate the ideas presented so far, consider a scenario where an AI system is used to make binary decisions.  The expected energy associated with the AI system is
\begin{equation}
\mathrm{E}[E]=E_{D}+AE_{X}+(1-A)\gamma E_{X}
\end{equation}

\begin{figure}[!t]
    \centering
    \includegraphics[width=1.04\columnwidth]{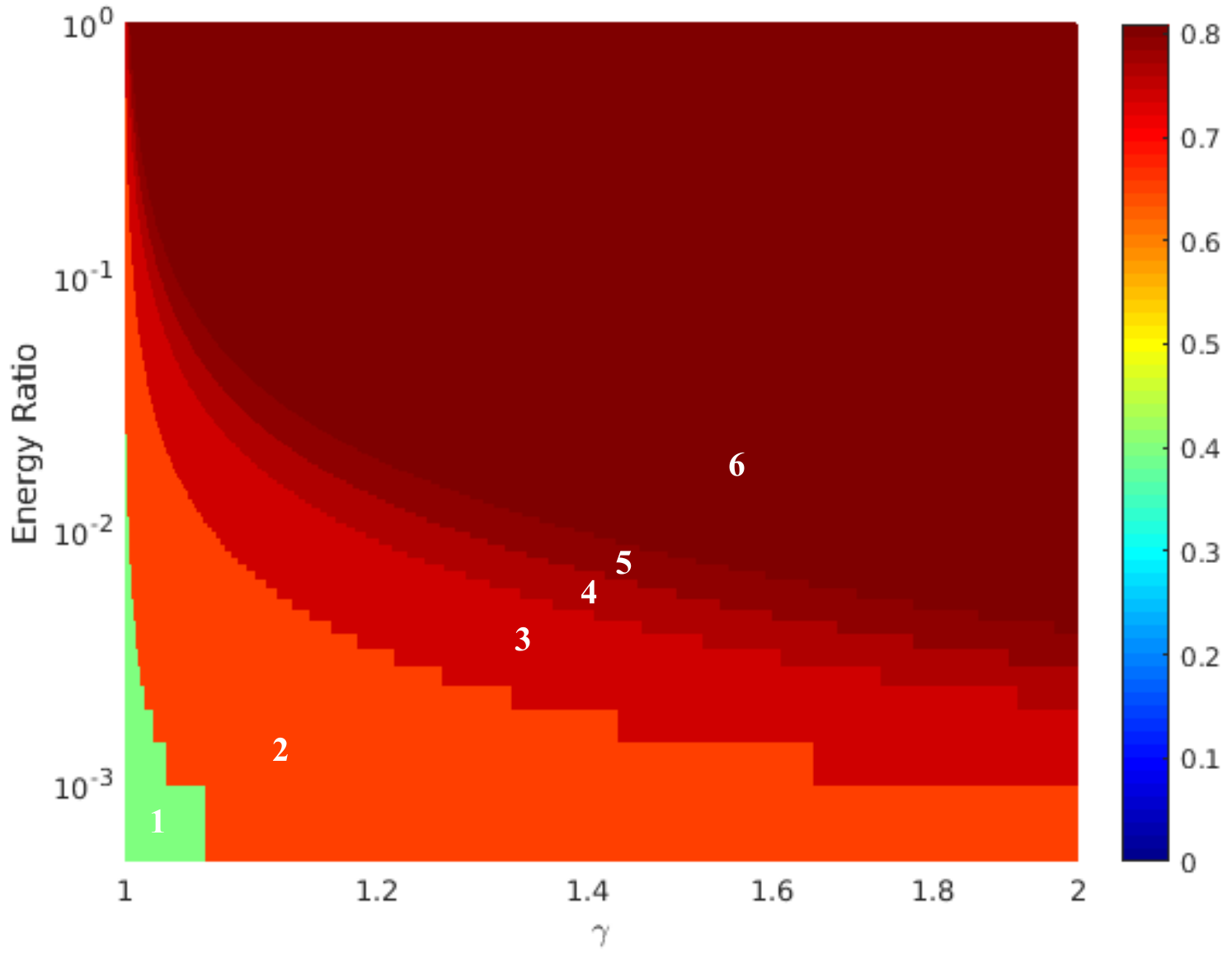}
    \caption{Optimal accuracy vs. energy ratio and $\gamma$ for a binary decision.  Numbers 1-6 indicate the number of CNN layers corresponding to the optimal accuracy.}
    \label{fig:optaccbinary}
\end{figure}

\noindent where $E_{D}$ and $E_{X}$ are the energies for making and executing a good decision, respectively.  The $\gamma$ coefficient is greater than or equal to one and indicates how much more energy is required to execute a bad decision.  This could correspond, for example, to the problem of an autonomous vehicle deciding to turn right or left at a fork based on a landmark.  If a correct turn is made, then the vehicle will reach the destination faster and use less energy.  If a wrong turn is made due to misclassification of the landmark, then the vehicle will need to travel further, using more energy.  With $E_{D}=E_{D_{max}}C_{D}$, optimizing the energy leads to 
\begin{equation}
A^{*}=\frac{1}{b}\mathrm{log}\left(\frac{(\gamma-1)E_{X}}{E_{D_{max}}ab}\right).
\label{eqn:optaccbinary}
\end{equation}
Comparing (\ref{eqn:optaccbinary}) with (\ref{eqn:acccostmin}) shows that the relative importance factor is $\alpha=(1+(\gamma-1)E_{X}/E_{D_{max}})^{-1}$.  So, when the ratio $E_{X}/E_{D_{max}}$ or $\gamma$ are small, the most important consideration is how much energy is required to make a decision.  As $\gamma$ and the energy ratio become larger, more emphasis is placed on having decisions based on accurate inference.  Figure \ref{fig:optaccbinary} shows the optimal accuracy from (\ref{eqn:optaccbinary}), rounded to the nearest achievable accuracy at a given number of layers (indicated in each region of the plot).  So, for this energy-accuracy curve shape, energy is minimized by using fewer than the maximum 6 layers if the energy ratio $E_{X}/E_{D_{min}}$ is well below 1 and $\gamma$ isn't too large.  Of course, for several implementations of an AI system, the energy ratio and $\gamma$ will both be large, leading to operation at the maximum accuracy (6 layers in this case).  However, for severely energy-constrained AI at the edge, it may often be necessary to operate at less than maximum accuracy in order to optimize the energy usage or otherwise-defined system cost.

\section{Dynamic Relative Importance Factor $\alpha$}
\label{section:dynamic}

\begin{figure}[!t]
\centering
\subfigure[]{
\centering
\includegraphics[width=0.4\columnwidth]{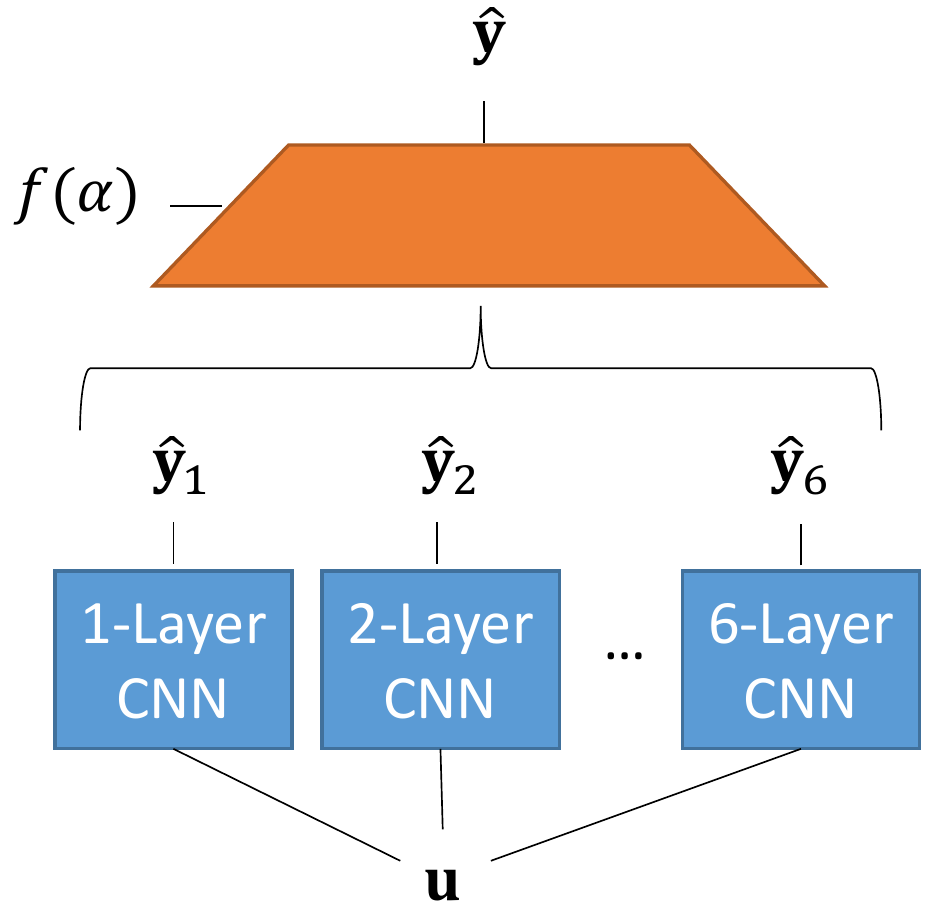}
}
\subfigure[]{
\centering
\includegraphics[width=0.75\columnwidth]{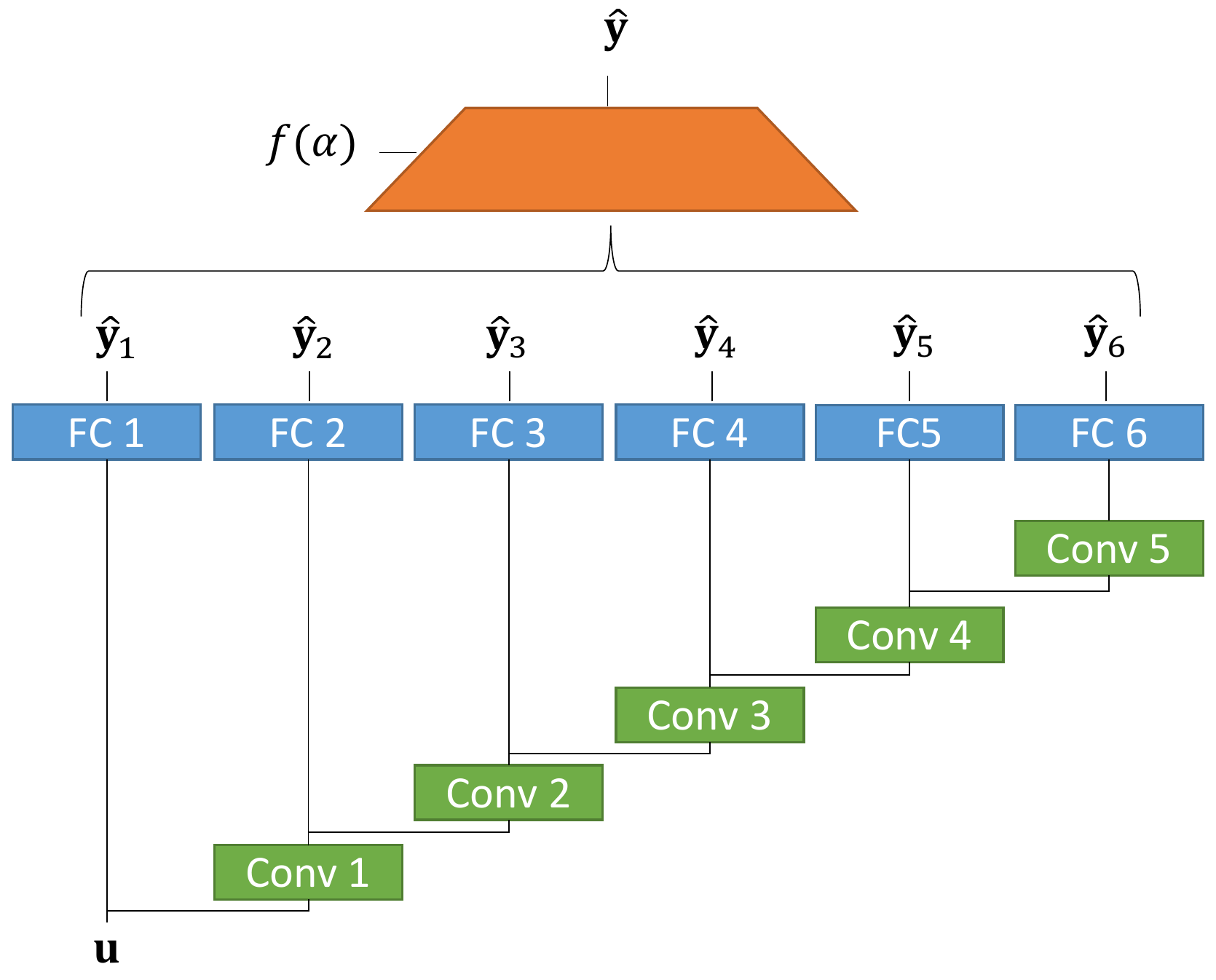}
}
\caption{(a) System that multiplexes between 6 CNNs depending on the relative importance factor of the cost (2.6 M parameters).  (b) System that multiplexes between the output layer of a single CNN based on the relative importance factor of the cost (2.4 M parameters).}
\label{fig:netconfigs}
\end{figure}

\begin{figure}[!t]
    \centering
    \includegraphics[width=\columnwidth]{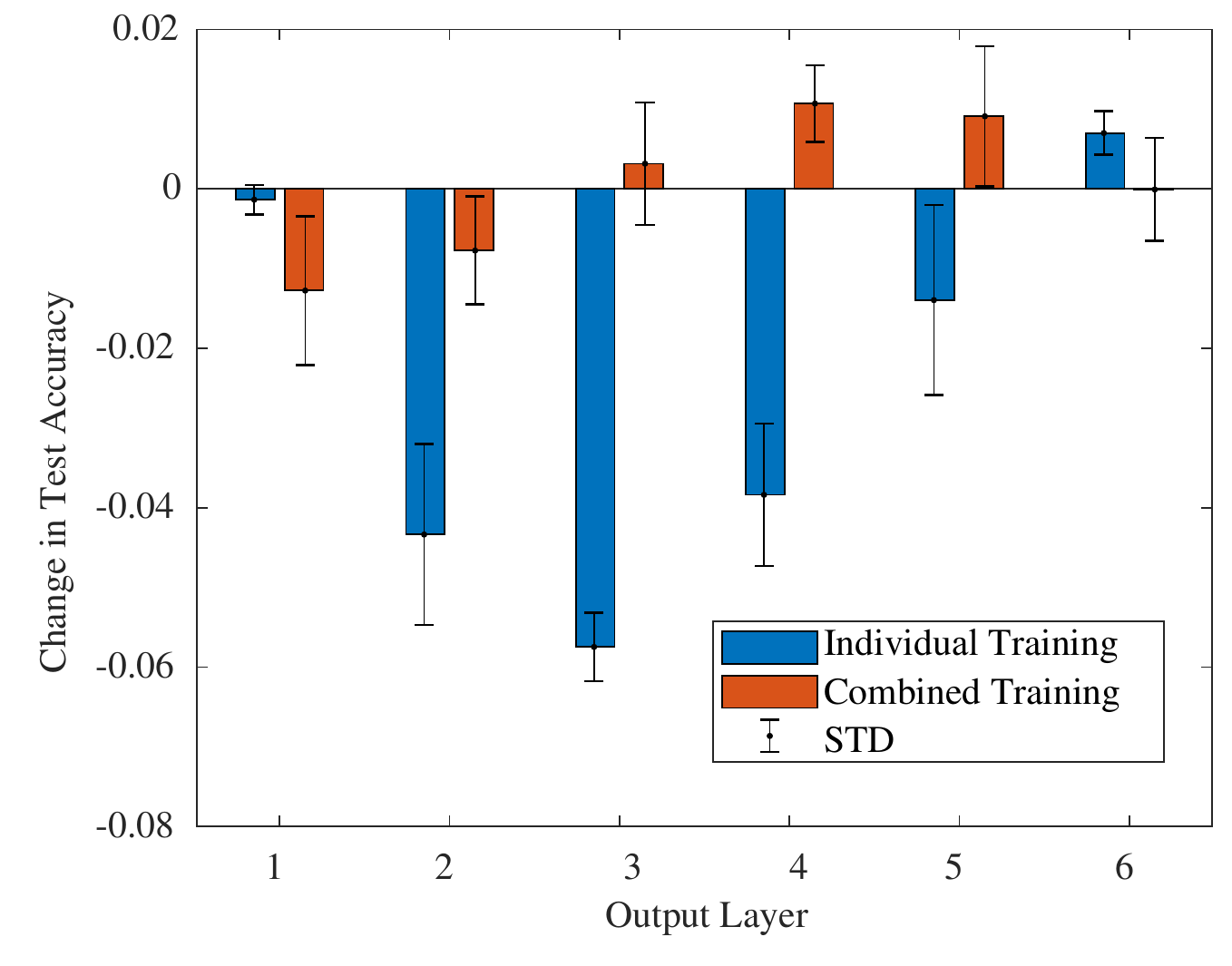}
    \caption{Change in test accuracy when 6 CNNs are combined into one multi-output CNN (Figure \ref{fig:netconfigs}(a) to Figure \ref{fig:netconfigs}(b)).  Individual training refers to the case where a 6-layer CNN is trained by minimizing the loss at the output of the last layer (phase 1) and then fully-connected layers are trained on the outputs of all of the 5 remaining layers (phase 2).  In this case, the convolutional layer weights are fixed in phase 2.  In combined training, all of the fully-connected softmax layers are trained concurrently by combining their cross entropy loss functions.}
    \label{fig:deltaacc}
\end{figure}

\begin{figure*}[!t]
\centering
\includegraphics[width=\textwidth]{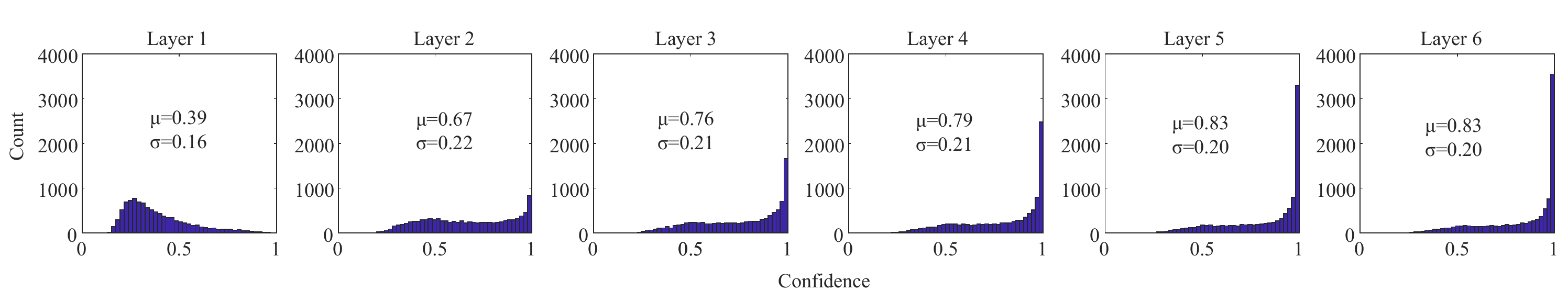}
\caption{Distributions of test set confidences in different layers of a 6-layer CNN.}
\label{fig:confdist}
\end{figure*} 

\begin{figure}
    \centering
    \includegraphics[width=\columnwidth]{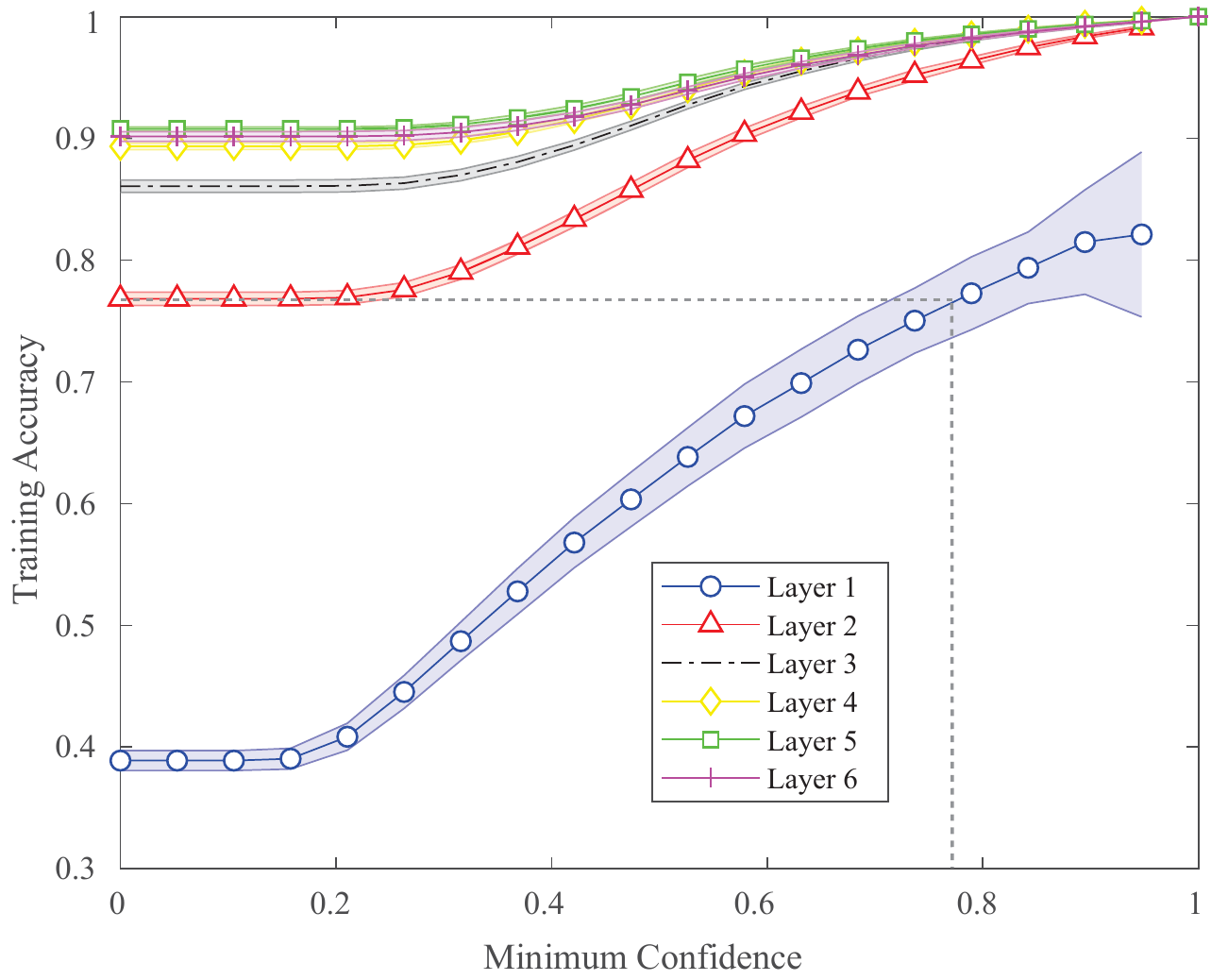}
    \caption{Training accuracy vs. minimum confidence for each of 6 CNN layers.}
    \label{fig:trainaccvsconf}
\end{figure}

 In the simplest scenario, $\alpha$ is fixed, allowing a network architecture to be chosen ahead of time based on the application constraints.  However, we envision a more complicated situation where $\alpha$ is not time-invariant and depends on a dynamic environment where the amount of energy available for computation and the consequences of bad  decisions are constantly changing.  In this case, it is critical that the system not only optimize resource usage based on $\alpha$ but also minimize the average amount of unused resources.  If we continue with the 6-layer CNN example discussed in the previous sections, a dynamic $\alpha$ value implies that the system should be able to make classifications based on the outputs of any of the 6 layers at any point in time.  One way to achieve this would be to multiplex between 6 different networks as shown in Figure \ref{fig:netconfigs}(a), where the selection is based on a function of the importance factor.  In an attempt to reduce the size of the implementation (number of parameters, silicon area, etc.), one may instead consider the structure shown in Figure \ref{fig:netconfigs}(b).  Here, a fully-connected softmax layer is connected to the output of each of 5 convolutional layers as well as the input.  This allows classifications to be made based on one of 6 sub-networks, each with an additional convolutional layer.  There is a similarity with the skip connections that are used in residual networks \cite{he2016deep}, except here only the information from lower layers is being used for the classification instead of combining it with information from higher layers.  Figure \ref{fig:deltaacc} shows the change in the test accuracy when moving from the structure in Figure \ref{fig:netconfigs}(a) to Figure \ref{fig:netconfigs}(b).  Individual training refers to the case where the 6-layer CNN in Figure \ref{fig:netconfigs}(b) is trained by minimizing the loss at the output of the last layer (phase 1) and then fully-connected layers are trained on the outputs of all of the 5 remaining layers (phase 2).  The convolutional layer weights are fixed in phase 2.  In combined training, all of the fully-connected softmax layers are trained concurrently by defining the total loss as the sum of their individual cross entropy losses.  It is observed that combining the loss functions yields better performance, with the layer-wise average change in accuracy for individual training $\approx -2.5\%$ and the layer-wise average change in accuracy for combined training $\approx 0.0\%$.  In fact, for some of the layers, the combined training approach gives better than baseline accuracy.  This is likely due to the combined loss function having a regularization effect which reduces overfitting.  In summary, the network structure in Figure \ref{fig:netconfigs}(b) allows the AI system to choose the number of layers that will be used for classification dynamically based on the relative cost factor without any change in accuracy over the baseline (Figure \ref{fig:netconfigs}(a)).  Futhermore, since the combined topology eliminates several redundant convolutional layers, there are $\approx 8\%$ fewer parameters.

\section{Reducing Energy Using Prediction Confidence and Dynamic \# of Layers}
\label{section:reducing}

This section explores the possibility of changing the baseline energy-accuracy curve shown in Figure \ref{fig:energy_accuracy} by leveraging the confidence of each layer's output.  Figure \ref{fig:confdist} shows the distribution of one run of test set confidences at each layer.  The confidence is defined as the largest value of each softmax output and is observed to increase on average in deeper layers.  However, there are some small number of test examples that result in high confidence predictions even in lower layers.  It may be possible to classify these examples with higher accuracy than the mean accuracy of the layer.  To explore this idea further, we investigated the accuracy of all training examples classified at a minimum confidence level for each layer (Figure \ref{fig:trainaccvsconf}).  The results show an approximately monotonically increasing training accuracy with increasing minimum confidence.  This means that, for example, if a minimum accuracy of $\approx 76\%$ is required by an application, then any inputs that correspond to a confidence of at least 0.76 in the first layer could be classified by the first layer (see dotted lines in Figure \ref{fig:trainaccvsconf}).  Otherwise, the second layer or higher layers would need to be used.  More generally, to classify an input, we can use the first layer that has a confidence level greater than or equal to a confidence level corresponding to the minimum desired accuracy.  The approach used is as follows:  For a given layer, if the layer's mean accuracy, regardless of confidence level, is greater than or equal to the desired accuracy then that layer is used as the output.  Otherwise, if the confidence of the prediction is above a specified threshold, then the layer will be used as the output.  If neither condition is met, then the data is further processed by the next layer.  If the input propagates to the final layer, then that layer is used for classification regardless of the confidence.  Or, formally, we are looking for the $l$ that satisfies:
\begin{equation}
\begin{aligned}
& {\text{minimize}}
& & l\in\{1,2,\ldots,L\} \\
& \text{subject to}
& & \beta_{1}A_{train_{l}}\ge \Tilde{A} \lor C_{l}\ge\beta_{2}\Tilde{A},
\end{aligned}
\label{eqn:opt}
\end{equation}
\noindent
where $A_{train_{l}}$ is the training set accuracy for layer $l$, $\Tilde{A}$ is the desired accuracy, and $C_{l}$ is the confidence level of layer $l$'s prediction.  The two $\beta$ coefficients are hyperparameters that account for the fact that the test accuracies will be lower than the training accuracies and accuracy at a given confidence level is generally less than the confidence level.  Note also that if (\ref{eqn:opt}) has no solution, then we set $l=L$  

\begin{figure}[!t]
\centering
\includegraphics[width=\columnwidth]{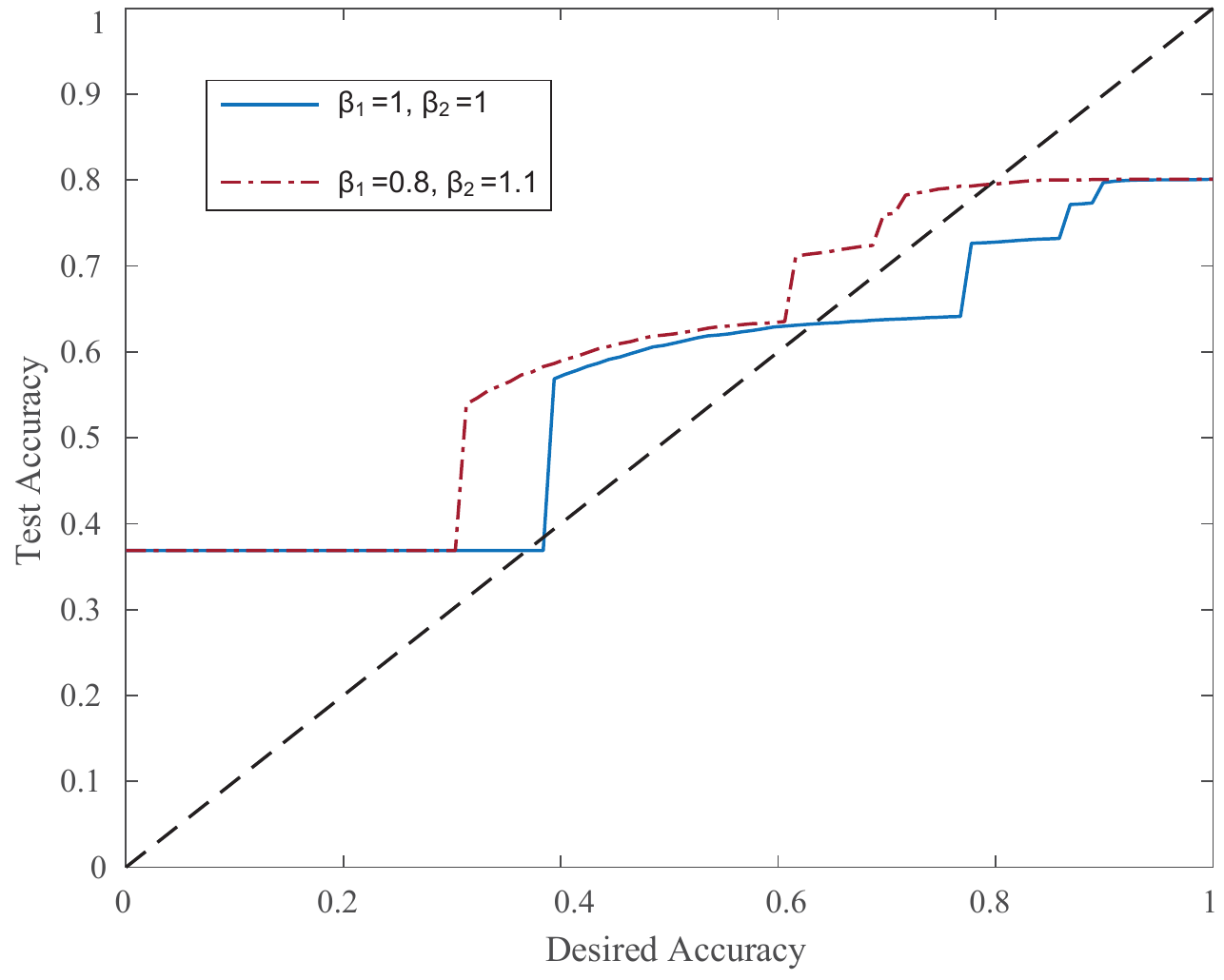}
\caption{Test accuracy vs. desired accuracy level for classification based on two different sets of $\beta$ values in (\ref{eqn:opt}).}
\label{fig:testvsdesired}
\end{figure}

\begin{figure}[!t]
\centering
\includegraphics[width=\columnwidth]{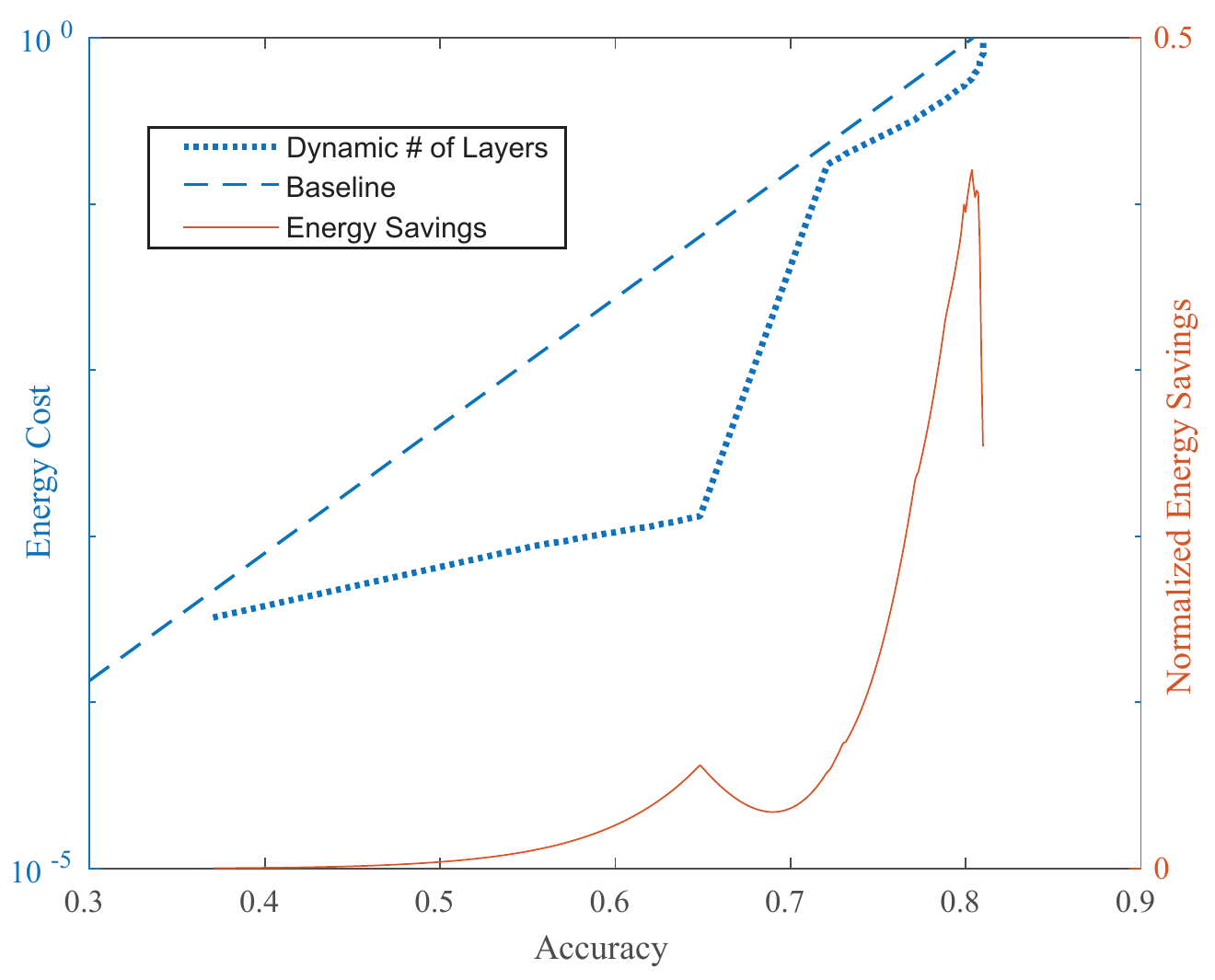}
\caption{Energy cost vs. accuracy for the 6-layer CNN implementation with dynamic \# of layers used for classification based on minimum desired accuracy.  The right axis also shows the energy savings over the baseline implementation (fixed  \# of layers used) normalized to the maximum energy cost.}
\label{fig:energyvsaccnew}
\end{figure}

The test accuracy was calculated vs. different levels of desired accuracy using the proposed dynamic layer scheme as shown in Figure \ref{fig:testvsdesired}.  For $\beta_{1}=\beta_{2}=\text{1}$, the test accuracy dips below the desired accuracy when the desired accuracy becomes large.  This can be remedied by modifying the $\beta$ hyperparameters such that a layer's mean accuracy or its confidence level has to be larger for that layer to be used for the final classification.  For example, Figure \ref{fig:testvsdesired} also shows the results for $\beta_{1}=0.8$ and $\beta_{2}=1.1$ which moves the majority of the curve above the $y=x$ line, until the mean accuracy of the full 6-layer network is reached at $\approx$80\% test accuracy.  Figure \ref{fig:energyvsaccnew} shows the energy vs. accuracy of the proposed dynamic layer scheme.  The baseline (from Figure \ref{fig:energy_accuracy}) is shown for comparison.  It can be observed that using a dynamic number of layers based on each layer's confidence yields a reduced energy cost at all accuracy levels.  In some cases, especially when the accuracy point is close to the network's maximum accuracy, the savings in energy over the baseline is significant (maximum savings of $\approx$42\%).  This gives an exciting possibility of dramatically improving the performance of energy-constrained AI systems at the edge by leveraging the high-confidence outputs of smaller sub-networks.
 
\section{Conclusions and Future Work}
\label{section:conclusions}

This paper provides some preliminary work on exploring the tradeoffs between energy and accuracy in AI hardware, especially for AI systems that are targeting edge application spaces (e.g. neuromorphic systems).  A framework was developed with a simple function to express the total cost of using an AI system as the weighted sum of two components:  the cost of the AI's decision making process and the cost of executing the decision.  Then, the specific case of energy-based cost was evaluated within that framework to explore the design space of CNNs with different numbers of convolutional layers.  Using MACs as a proxy for energy, we showed that the optimal number of layers in a CNN performing a binary decision can be easily calculated based on the energy-accuracy curve, the maximum decision energy, and the energy required to execute the decision.  It was also demonstrated that each layer of a multilayer CNN can be used to classify an input with confidence and accuracy increasing in deeper layers.  In addition, using high-confidence predictions in earlier layers of the network enables a significant reduction (up to 42\%) in the network's energy.  Potential avenues for future work include further exploration of using a dynamic number of CNN filters, combining each layer's prediction with the previous layers' for better accuracy, and experimentation with circuit-level (e.g. SPICE) simulations to achieve better energy modeling.




\bibliographystyle{IEEEtran}
\bibliography{refs.bib}

\end{document}